\documentclass{article}
\usepackage{spconf,amsmath,graphicx,hyperref}
\usepackage{makecell}
\makeatletter
\let\oldshortstack\shortstack
\renewcommand{\shortstack}[2][c]{%
  \setbox\z@\hbox{\oldshortstack[#1]{#2}}%
  \rule{\z@}{\dimexpr\ht\z@+0.24pt\relax}\box\z@}
\makeatother
\usepackage{amsfonts}
\usepackage{multirow}      
\usepackage[table]{xcolor} 
\definecolor{summaryrow}{RGB}{220, 230, 242}
\definecolor{beatgreen}{RGB}{198, 232, 210}

\newcommand{\secondbest}[1]{%
  \begingroup
  \setbox0=\hbox{#1}%
  \rlap{\raisebox{-1.2pt}[0pt][0pt]{\rule{\wd0}{0.35pt}}}%
  \box0\endgroup}

\newcommand{\gpos}[1]{\colorbox{beatgreen}{\strut$+$#1}}
\newcommand{\gnil}[1]{\strut#1}

\title{SleepPairNet: Learning Multimodal Pediatric Sleep Representations with Random-Pair Cross-Attention}
\name{Author(s) Name(s)\thanks{Thanks to XYZ agency for funding.}}
\address{Author Affiliation(s)}
\twoauthors
  {Saurav Raj Pandey}
  {Department of Computer Science\\
   University of North Carolina at Chapel Hill\\
   Chapel Hill, NC, USA}
  {Harlin Lee}
  {School of Data and Information Sciences\\
   University of North Carolina at Chapel Hill\\
   Chapel Hill, NC, USA}

\begin{document}


\raggedbottom

\maketitle

\begin{abstract}
Analyzing physiological sleep signals is important
for assessing sleep quality and detecting
sleep-related disorders. Learning inter-modal
relationships across physiological signals can
provide richer representations for downstream
sleep analysis. We introduce \textbf{SleepPairNet}, a multimodal self-supervised learning (SSL) framework that learns pairwise interactions between physiological signals
through cross-attention. We propose
a random modality-pair selection strategy that
exposes the model to diverse signal pairs during
pretraining, enabling flexibility across modality
combinations at test time. Pretrained on the NCH dataset, SleepPairNet outperforms the evaluated SSL baselines on all respiratory tasks while remaining competitive on EEG-related tasks, and shows even stronger results when transferred to the external CHAT dataset.

\end{abstract}

\section{Introduction}
Sleep is crucial to the physical and mental well-being of children, with poor sleep associated with adverse outcomes such as anxiety, depression, and hypertension
\cite{sleep_disruption}. In clinical practice, polysomnography (PSG) records multiple physiological signals (EEG, EOG, EMG, etc.) overnight, but its analysis requires costly expert annotations of sleep stages and events. Self-supervised learning (SSL) can leverage unlabeled PSG data to learn representations
transferable to downstream tasks with limited
labeled data, reducing annotation requirements and improving generalization.

Recent sleep SSL methods have increasingly incorporated multiple physiological signals and explored flexible designs for varying channel configurations. Cross-attention provides a natural mechanism for learning relationships between modalities by allowing representations from one physiological signal to directly attend to those of another, but remains relatively underexplored in sleep SSL.

We propose \textbf{SleepPairNet}, a multimodal SSL
framework for pediatric PSG that learns cross-modal
representations through cross-attention. We introduce a random modality-pair strategy that samples a new signal
pair at each training mini-batch, exposing a single model to diverse modality
combinations. Pretrained on NCH, SleepPairNet
outperforms the evaluated SSL baselines on
respiratory-related tasks (apnea, hypopnea, and oxygen desaturation detection) while remaining close on EEG-related tasks (sleep scoring and EEG arousal detection). When transferred to CHAT without further SSL or backbone fine-tuning, it achieves even larger respiratory-task gains and stronger relative performance on EEG-related tasks.

\begin{figure}[!t]
    \includegraphics[width=\columnwidth]{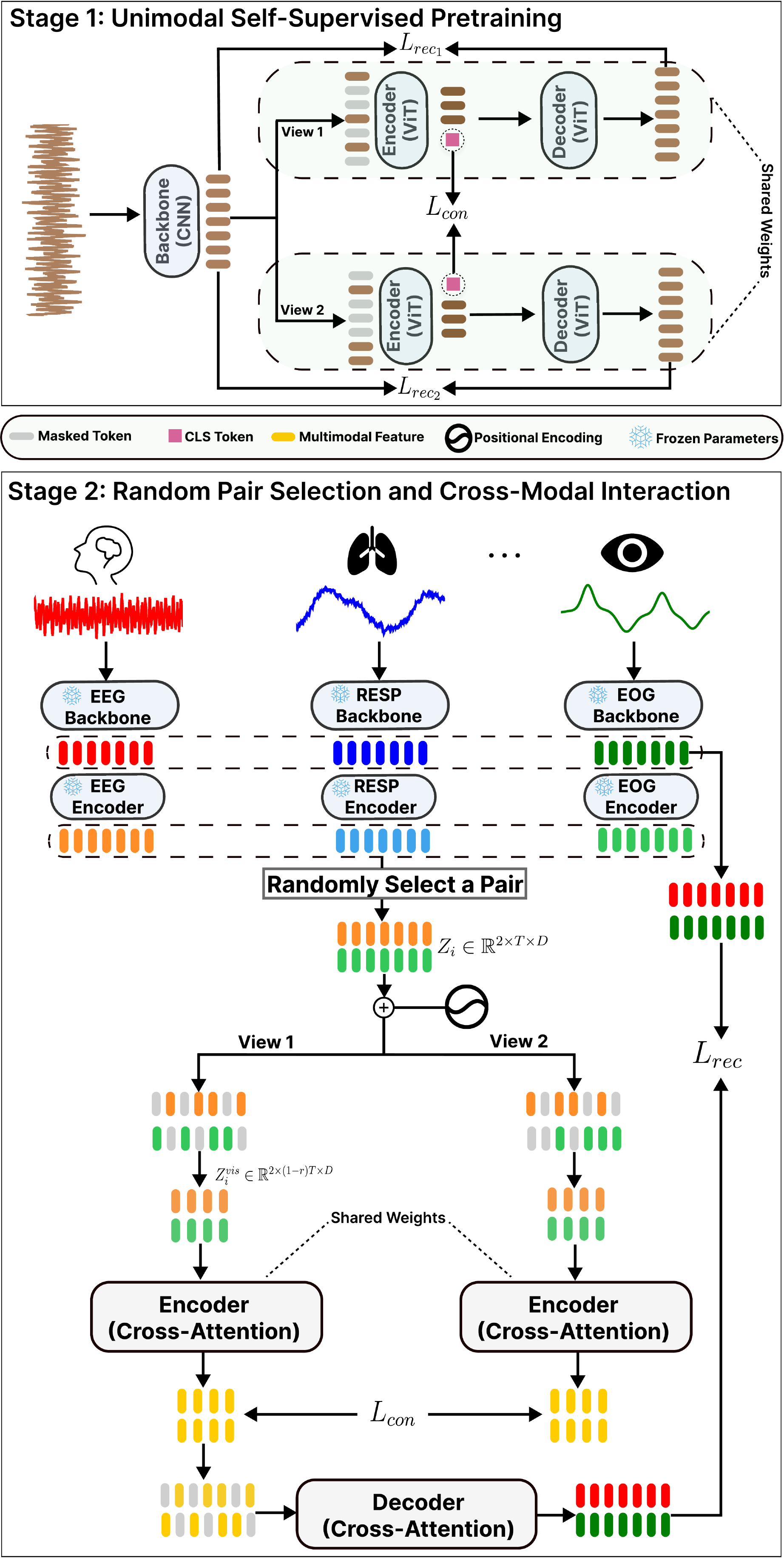}
    \caption{Overview of SleepPairNet's architecture.}
    \label{fig:sleeppairnet}
\end{figure}

\section{Related Work}
Sleep SSL has largely focused on learning representations from unimodal EEG, including BENDR \cite{kostas2021bendr}, EEGPT \cite{wang2024eegpt}, and NeuroNet \cite{lee2024neuronet}. More recent approaches have extended SSL to multimodal physiological signals, including COCOA \cite{deldari2022cocoa}, SleepFM \cite{thapa2026multimodal}, SynthSleepNet \cite{synthsleepnet}, and PedSleepMAE \cite{pandey2024pedsleepmae}. SleepFM is channel-agnostic and handles variability in channel number via attention pooling, while SynthSleepNet and PedSleepMAE assume fixed input modalities.

Cross-attention has been used to model EEG--EOG
interactions in supervised sleep scoring
\cite{CASleepNet, gui2026mfsleepnet}. Outside sleep, CardioFM
\cite{hassanuzzaman2026cardiofm} and MultiSSL
\cite{xue2025multimodal} explore cross-modal
attention in SSL but use fixed sets of two to three
modalities. In contrast, SleepPairNet applies
cross-attention to randomly sampled
physiological signal pairs during SSL, enabling one
model to learn across diverse modality combinations.

Random channel selection has also been used in supervised sleep models:
U-Sleep~\cite{UNet} samples random EEG/EOG channels for montage robustness,
and wav2sleep~\cite{carter2024wav2sleep} randomly drops signals to handle
variable inputs. SleepPairNet brings random selection to SSL, sampling
heterogeneous modality pairs to learn transferable cross-modal relationships.

\section{Methodology}
We describe our datasets and SleepPairNet's training strategy. All code will be made public upon acceptance of the paper.

\subsection{Dataset and Preprocessing}
We pretrain SleepPairNet on the publicly available Nationwide Children's Hospital (NCH) Sleep DataBank, a large pediatric PSG dataset collected in clinical settings \cite{lee2022large}. We use 2{,}379 overnight PSGs containing 16 channels spanning EEG, EOG, EMG, respiratory, and other physiological signals.

For cross-dataset evaluation, we use 422 baseline
PSGs from the Childhood Adenotonsillectomy Trial
(CHAT) \cite{zhang2018national,
marcus2013randomized} dataset, restricted to 12 channels
overlapping with NCH and used only to evaluate
NCH-pretrained representations.

Both datasets provide 30-second annotations for sleep stages (Wake, N1, N2, N3, and REM) and clinically relevant events, including apnea, hypopnea, EEG arousal, and oxygen desaturation. We treat each non-overlapping 30-second epoch as an individual sample, standardize each channel to zero mean and unit variance, and use patient-level 80/10/10 train/validation/test splits.

\subsection{SleepPairNet Architecture}
As shown in Figure~\ref{fig:sleeppairnet}, SleepPairNet treats each PSG channel as a modality and uses two-stage pretraining: (1) unimodal SSL per channel and (2) bimodal SSL on randomly sampled modality pairs.

\subsubsection{Stage 1: Unimodal Self-Supervised Pretraining}
\label{sec:stage1}
We independently pretrain a separate NeuroNet \cite{lee2024neuronet} for each of the 16 physiological channels, retaining its original architecture and SSL objectives while adapting preprocessing and training configurations to our experimental setting. Briefly, a CNN backbone produces frame-level features to which positional encodings are added. Two views are then generated using independent random masking, with $75\%$ of the tokens masked in each view. The visible tokens from each view are passed through a Vision Transformer (ViT) encoder, while a smaller Transformer decoder reconstructs the corresponding CNN backbone features at the masked positions. The CLS tokens from both views are projected into a contrastive space and trained with the NT-Xent loss \cite{chen2020simple}.

The training objective is
\begin{equation}
L_{\mathrm{tot}}
=
L_{\mathrm{rec}_1}
+
L_{\mathrm{rec}_2}
+
\alpha L_{\mathrm{con}},
\end{equation}
where $L_{\mathrm{rec}_1}$ and
$L_{\mathrm{rec}_2}$ are masked reconstruction
losses, $L_{\mathrm{con}}$ is the NT-Xent loss,
and $\alpha$ controls its contribution.

\subsubsection{Stage 2: Random Pair Selection and Cross-Modal Interaction}
\label{sec:stage_2}
After Stage 1, each channel's pretrained CNN backbone and encoder are frozen and used to obtain
$h_m\in\mathbb{R}^{(T+1)\times D}$ for each modality $m$, where $T$ and
$D$ denote the number of temporal tokens and
embedding dimension, respectively. At each iteration, we randomly
select two of the 16 channels, discard their CLS
tokens, and stack the temporal tokens of each sample $i$ to form
$Z_i\in\mathbb{R}^{2\times T\times D}$.

After adding positional encodings to $Z_i$, we independently
mask each modality at masking ratio $r=0.5$, yielding
$Z_i^{\mathrm{vis}} \in\mathbb{R}^{2\times(1-r)T\times D}$. Then, $Z_i^{\mathrm{vis}}$ is processed by a cross-attention Transformer encoder, where the tokens of each modality serve as queries and attend to the keys and values derived exclusively from the other modality. A cross-attention Transformer decoder restores the masked positions using learned mask tokens and reconstructs the corresponding frame-level CNN features. Reconstruction loss is computed only at masked positions.

For contrastive learning, we generate two independently masked views of the same modality pair. For each view, we average the encoder representations across both modalities and temporal tokens to obtain a
single multimodal representation, which is projected into a contrastive space and trained with the NT-Xent loss. Reconstruction is performed on only one view, giving the overall training objective
$L_{\mathrm{rec}} + L_{\mathrm{con}}$.

\subsection{Downstream Classification}

\subsubsection{Linear Probing and Channel Selection}
We evaluate representations before and after SleepPairNet's Stage-2 cross-modal pretraining. Given the stacked Stage-1
representations $Z_i \in \mathbb{R}^{2 \times T \times D}$ defined in
Section~\ref{sec:stage_2}, our ``Concat'' ablation mean-pools each modality
over the temporal dimension to obtain two $D$-dimensional vectors, which are concatenated and provided as input to a linear classifier for each downstream task.

For SleepPairNet, we set $r=0$, such that $Z_i^{\mathrm{vis}}=Z_i$, and pass $Z_i$ through the cross-attention encoder.
We then apply the same temporal mean-pooling,
concatenation, and linear classification as in Concat. This comparison isolates the
contribution of Stage-2 cross-modal pretraining.

For each task, we use individual-channel rankings from \cite{pandey2026bitimecrossnet}, obtained by linear probing on a subset of the NCH train split, as a starting point for selecting modality pairs. We evaluate combinations of highly ranked channels and additional task-relevant pairs identified using domain knowledge (e.g., SpO$_2$ and  CAPNO for respiratory events), selecting the pair with the highest SleepPairNet validation F1 for each task.

For NCH, all representation models are frozen after SSL and only the linear classifiers are trained for each task. For CHAT, NCH-pretrained representations are transferred directly without further self-supervised training or backbone fine-tuning, and only linear classifiers are trained. Class-weighted cross-entropy is used to address class imbalance.

\subsubsection{Baselines}
All representation models are pretrained on the same NCH train split. Aside from our ``Concat'' ablation, we compare against SleepFM, PedSleepMAE, SynthSleepNet, and NeuroNet. SleepFM and PedSleepMAE are evaluated using the
same task-specific modality pairs as SleepPairNet, so that all multimodal methods receive identical inputs, while unimodal NeuroNet is evaluated separately on each modality of the pair. Since SleepFM does not
specify a linear-probing procedure, we follow
\cite{pandey2026bitimecrossnet} by mean-pooling its temporal embeddings separately for each modality and concatenating the resulting representations. For PedSleepMAE, we follow the authors' original linear-probing implementation.

For SynthSleepNet, we restrict evaluation to modality configurations studied in the original work \cite{synthsleepnet}, rather than extending the framework
to previously unevaluated respiratory modalities. EEG, EOG, and EMG in our
SynthSleepNet experiments correspond to EEG C4--M1, EOG LOC--M2, and EMG
CHIN1--CHIN2, respectively.

\definecolor{summaryrow}{RGB}{220, 230, 242}
\definecolor{beatgreen}{RGB}{198, 232, 210}

\newcommand{\upcell}[1]{\raisebox{0.7ex}{#1}}

\begin{table*}[!t]
\centering
\caption{NCH linear-probing results
(mean (SD) over linear-probe seeds 41--43).
Bold/underline: best/second-best.
Parentheses under task name: binary-task prevalence.
\colorbox{summaryrow}{\strut Shaded}:
$\Delta$F1/$\Delta$AUPRC of best Ours vs.\
best non-Ours;
\colorbox{beatgreen}{\strut green} = gain.}
\label{tab:nch_baseline_comparison}
\fontsize{9}{10}\selectfont
\setlength{\tabcolsep}{1.13pt}
\renewcommand{\arraystretch}{1.05}

\begin{tabular}{@{}l|l|cccc|cccc|cccc|cccc|cccc@{}}
\hline

\multirow{3}{*}{\textbf{Model}} &
\multirow{3}{*}{\textbf{Modalities}} &
\multicolumn{20}{c}{\textbf{Performance (\%)}} \\

\cline{3-22}

& &
\multicolumn{4}{c|}{\textbf{Sleep Scoring}} &
\multicolumn{4}{c|}{\shortstack{\textbf{Apnea}\\(0.8\%)}} &
\multicolumn{4}{c|}{\shortstack{\textbf{EEG Arousal}\\(4.6\%)}} &
\multicolumn{4}{c|}{\shortstack{\textbf{Oxygen Desat.}\\(8.7\%)}} &
\multicolumn{4}{c}{\shortstack{\textbf{Hypopnea}\\(1.9\%)}} \\

\cline{3-22}

& &
Acc. & F1 & ROC & PR &
Acc. & F1 & ROC & PR &
Acc. & F1 & ROC & PR &
Acc. & F1 & ROC & PR &
Acc. & F1 & ROC & PR \\

\hline


SleepFM &
$c_1, c_2$ &
\shortstack{66.8\\(0.0)} &
\shortstack{60.1\\(0.1)} &
\shortstack{88.5\\(0.1)} &
\shortstack{64.7\\(0.1)} &

\shortstack{\secondbest{98.2}\\(0.0)} &
\shortstack{14.8\\(0.2)} &
\shortstack{80.4\\(0.1)} &
\shortstack{7.3\\(0.2)} &

\shortstack{92.8\\(0.1)} &
\shortstack{\textbf{41.7}\\(0.2)} &
\shortstack{\textbf{87.4}\\(0.1)} &
\shortstack{\textbf{36.0}\\(0.5)} &

\shortstack{84.8\\(0.3)} &
\shortstack{36.2\\(0.1)} &
\shortstack{77.8\\(0.1)} &
\shortstack{31.9\\(0.1)} &

\shortstack{{96.3}\\(0.1)} &
\shortstack{17.5\\(0.3)} &
\shortstack{77.8\\(0.1)} &
\shortstack{10.7\\(0.1)}
\\

\hline


PedSleepMAE &
$c_1, c_2$ &
\shortstack{60.5\\(1.0)} &
\shortstack{46.2\\(1.5)} &
\shortstack{84.1\\(0.1)} &
\shortstack{53.5\\(0.2)} &
\shortstack{97.8\\(0.2)} &
\shortstack{11.4\\(0.1)} &
\shortstack{74.1\\(0.1)} &
\shortstack{4.4\\(0.0)} &
\shortstack{87.4\\(0.6)} &
\shortstack{24.4\\(0.1)} &
\shortstack{75.3\\(0.0)} &
\shortstack{13.9\\(0.0)} &
\shortstack{81.3\\(0.2)} &
\shortstack{27.0\\(0.0)} &
\shortstack{69.7\\(0.0)} &
\shortstack{21.1\\(0.0)} &
\shortstack{94.0\\(0.1)} &
\shortstack{14.9\\(0.1)} &
\shortstack{73.0\\(0.2)} &
\shortstack{7.2\\(0.0)} 
\\

\hline


\multirow{2}{*}{SynthSleepNet} &
\makecell[l]{EEG, EOG} &
\shortstack{\textbf{74.8}\\(0.3)} &
\shortstack{\textbf{67.2}\\(0.2)} &
\shortstack{\textbf{92.5}\\(0.0)} &
\shortstack{\textbf{72.2}\\(0.0)} &

\shortstack{92.2\\(5.5)} &
\shortstack{5.5\\(0.4)} &
\shortstack{73.6\\(1.1)} &
\shortstack{2.6\\(0.3)} &

\shortstack{\textbf{93.2}\\(0.1)} &
\shortstack{40.1\\(0.2)} &
\shortstack{84.3\\(0.3)} &
\shortstack{33.0\\(0.2)} &

\shortstack{71.3\\(0.9)} &
\shortstack{26.7\\(0.1)} &
\shortstack{70.9\\(0.1)} &
\shortstack{18.0\\(0.1)} &

\shortstack{93.3\\(0.3)} &
\shortstack{10.1\\(0.4)} &
\shortstack{73.2\\(0.4)} &
\shortstack{5.6\\(0.2)}
\\

\cline{2-22}

&
\makecell[l]{EEG, EMG} &
\shortstack{\secondbest{73.2}\\(0.3)} &
\shortstack{\secondbest{65.9}\\(0.2)} &
\shortstack{\secondbest{91.9}\\(0.0)} &
\shortstack{\secondbest{71.2}\\(0.1)} &

\shortstack{95.2\\(0.1)} &
\shortstack{8.3\\(0.2)} &
\shortstack{77.0\\(0.8)} &
\shortstack{3.7\\(0.2)} &

\shortstack{\secondbest{93.0}\\(0.2)} &
\shortstack{\secondbest{41.6}\\(0.2)} &
\shortstack{\secondbest{86.8}\\(0.1)} &
\shortstack{\secondbest{35.4}\\(0.1)} &

\shortstack{71.9\\(1.1)} &
\shortstack{27.1\\(0.2)} &
\shortstack{71.5\\(0.2)} &
\shortstack{18.6\\(0.1)} &

\shortstack{90.8\\(1.5)} &
\shortstack{11.7\\(1.0)} &
\shortstack{76.3\\(0.9)} &
\shortstack{6.5\\(0.6)}
\\

\hline

\multirow{2}{*}{NeuroNet} &
$c_1$ &
\shortstack{65.3\\(1.6)} &
\shortstack{56.8\\(0.4)} &
\shortstack{86.0\\(0.3)} &
\shortstack{61.3\\(0.3)} &
\shortstack{\textbf{98.3}\\(0.0)} &
\shortstack{\secondbest{19.7}\\(0.3)} &
\shortstack{83.8\\(0.1)} &
\shortstack{\secondbest{10.8}\\(0.0)} &
\shortstack{92.2\\(0.1)} &
\shortstack{29.4\\(0.7)} &
\shortstack{75.5\\(0.5)} &
\shortstack{20.2\\(0.7)} &
\shortstack{78.7\\(0.8)} &
\shortstack{29.5\\(1.2)} &
\shortstack{72.6\\(0.9)} &
\shortstack{20.5\\(0.7)} &
\shortstack{89.3\\(0.4)} &
\shortstack{6.8\\(1.1)} &
\shortstack{64.4\\(0.1)} &
\shortstack{3.7\\(0.4)}
\\

\cline{2-22}

&
$c_2$ &
\shortstack{64.6\\(1.3)} &
\shortstack{57.2\\(0.6)} &
\shortstack{86.8\\(0.3)} &
\shortstack{61.5\\(0.4)} &
\shortstack{94.1\\(3.7)} &
\shortstack{4.3\\(0.8)} &
\shortstack{67.6\\(2.9)} &
\shortstack{2.0\\(0.4)} &
\shortstack{91.1\\(0.1)} &
\shortstack{34.3\\(0.1)} &
\shortstack{84.4\\(0.3)} &
\shortstack{26.5\\(0.3)} &
\shortstack{\secondbest{87.2}\\(0.1)} &
\shortstack{40.3\\(0.0)} &
\shortstack{79.6\\(0.0)} &
\shortstack{36.1\\(0.1)} &
\shortstack{96.2\\(0.1)} &
\shortstack{24.4\\(0.1)} &
\shortstack{81.9\\(0.1)} &
\shortstack{15.3\\(0.1)}
\\

\hline
\hline

\multirow{2}{*}{\textbf{Ours}} &
\shortstack[l]{$c_1,c_2$\\(Concat)} &
\shortstack{72.0\\(1.6)} &
\shortstack{63.6\\(0.5)} &
\shortstack{90.2\\(0.2)} &
\shortstack{68.2\\(0.4)} &
\shortstack{97.9\\(0.2)} &
\shortstack{17.8\\(0.7)} &
\shortstack{\secondbest{84.3}\\(0.7)} &
\shortstack{9.6\\(0.5)} &
\shortstack{92.5\\(0.3)} &
\shortstack{37.7\\(0.9)} &
\shortstack{85.0\\(0.7)} &
\shortstack{30.5\\(0.9)} &
\shortstack{\secondbest{87.2}\\(0.0)} &
\shortstack{\secondbest{41.5}\\(0.6)} &
\shortstack{\secondbest{80.9}\\(0.4)} &
\shortstack{\secondbest{36.6}\\(0.8)} &
\shortstack{\textbf{96.8}\\(0.0)} &
\shortstack{\secondbest{25.4}\\(0.1)} &
\shortstack{\secondbest{83.4}\\(0.8)} &
\shortstack{\secondbest{17.8}\\(0.1)}
\\

\cline{2-22}

&
\shortstack[l]{$c_1,c_2$\\(SleepPairNet)} &
\shortstack{71.2\\(0.4)} &
\shortstack{63.4\\(0.3)} &
\shortstack{90.2\\(0.1)} &
\shortstack{68.2\\(0.2)} &
\shortstack{97.9\\(0.0)} &
\shortstack{\textbf{21.7}\\(0.1)} &
\shortstack{\textbf{87.3}\\(0.1)} &
\shortstack{\textbf{12.5}\\(0.0)} &
\shortstack{92.0\\(0.2)} &
\shortstack{36.7\\(0.3)} &
\shortstack{84.8\\(0.1)} &
\shortstack{29.3\\(0.4)} &
\shortstack{\textbf{88.3}\\(0.2)} &
\shortstack{\textbf{45.1}\\(0.2)} &
\shortstack{\textbf{82.9}\\(0.0)} &
\shortstack{\textbf{41.7}\\(0.4)} &
\shortstack{\secondbest{96.4}\\(0.4)} &
\shortstack{\textbf{30.2}\\(0.3)} &
\shortstack{\textbf{85.8}\\(0.1)} &
\shortstack{\textbf{21.1}\\(0.2)}
\\

\hline

\multicolumn{2}{l|}{
\cellcolor{summaryrow}
$\Delta$F1\,/\,$\Delta$AUPRC
} &
\multicolumn{4}{c|}{
\cellcolor{summaryrow}\gnil{$-$3.6}\,/\,\gnil{$-$4.0}
} &
\multicolumn{4}{c|}{
\cellcolor{summaryrow}\gpos{2.0}\,/\,\gpos{1.7}
} &
\multicolumn{4}{c|}{
\cellcolor{summaryrow}\gnil{$-$4.0}\,/\,\gnil{$-$5.5}
} &
\multicolumn{4}{c|}{
\cellcolor{summaryrow}\gpos{4.8}\,/\,\gpos{5.6}
} &
\multicolumn{4}{c}{
\cellcolor{summaryrow}\gpos{5.8}\,/\,\gpos{5.8}
}
\\
\hline
\end{tabular}

\parbox{\textwidth}{
\fontsize{9}{10.0}\selectfont
\textit{Note:} Task-specific $(c_1,c_2)$ pairs:
Sleep = (EEG C4--M1, EOG LOC--M2);
Apnea = (SpO$_2$, EOG ROC--M1);
Arousal = (EEG C4--M1, EMG CHIN1--CHIN2);
Desaturation = (CAPNO, SpO$_2$);
Hypopnea = (EEG CZ--O1, SpO$_2$).
Concat and SleepPairNet use mean-pooled Stage-1
and Stage-2 representations, respectively.
F1 is macro-F1 for sleep scoring and binary F1
otherwise; ROC/PR denote AUROC/AUPRC.}
\end{table*}

\begin{table*}[!t]
\centering

\caption{Evaluation of NCH-pretrained models on the external CHAT cohort. Layout and conventions as in Table \ref{tab:nch_baseline_comparison}.}
\label{tab:chat_comparison}

\fontsize{9.0}{10.0}\selectfont
\setlength{\tabcolsep}{1.13pt}
\renewcommand{\arraystretch}{1}
\begin{tabular}{@{}l|l|cccc|cccc|cccc|cccc|cccc@{}}
\hline

\multirow{3}{*}{\textbf{Model}} &
\multirow{3}{*}{\textbf{Modalities}} &
\multicolumn{20}{c}{\textbf{Performance (\%)}} \\

\cline{3-22}

& &
\multicolumn{4}{c|}{\textbf{Sleep Scoring}} &
\multicolumn{4}{c|}{\shortstack{\textbf{Apnea}\\(2.4\%)}} &
\multicolumn{4}{c|}{\shortstack{\textbf{EEG Arousal}\\(8.0\%)}} &
\multicolumn{4}{c|}{\shortstack{\textbf{Oxygen Desat.}\\(17.0\%)}} &
\multicolumn{4}{c}{\shortstack{\textbf{Hypopnea}\\(3.9\%)}} \\

\cline{3-22}

& &
Acc. & F1 & ROC & PR &
Acc. & F1 & ROC & PR &
Acc. & F1 & ROC & PR &
Acc. & F1 & ROC & PR &
Acc. & F1 & ROC & PR \\

\hline




SleepFM &
$c_1, c_2$ &
\shortstack{70.9\\(0.1)} &
\shortstack{65.5\\(0.2)} &
\shortstack{90.7\\(0.1)} &
\shortstack{68.9\\(0.1)} &
\shortstack{92.7\\(0.1)} &
\shortstack{10.7\\(0.1)} &
\shortstack{73.9\\(0.2)} &
\shortstack{5.4\\(0.1)} &
\shortstack{\secondbest{90.3}\\(0.1)} &
\shortstack{\secondbest{52.6}\\(0.1)} &
\shortstack{90.6\\(0.0)} &
\shortstack{\textbf{49.2}\\(0.1)} &
\shortstack{\secondbest{81.6}\\(0.0)} &
\shortstack{\secondbest{52.3}\\(0.0)} &
\shortstack{\secondbest{82.6}\\(0.1)} &
\shortstack{\secondbest{49.2}\\(0.1)} &
\shortstack{90.4\\(0.1)} &
\shortstack{19.7\\(0.0)} &
\shortstack{79.1\\(0.1)} &
\shortstack{12.4\\(0.1)}
\\

\hline



PedSleepMAE &
$c_1, c_2$ &
\shortstack{44.5\\(0.9)} &
\shortstack{38.1\\(0.5)} &
\shortstack{74.3\\(0.1)} &
\shortstack{39.1\\(0.2)} &
\shortstack{41.4\\(5.1)} &
\shortstack{4.9\\(0.2)} &
\shortstack{60.2\\(0.0)} &
\shortstack{3.0\\(0.0)} &
\shortstack{78.4\\(0.8)} &
\shortstack{27.1\\(0.1)} &
\shortstack{71.2\\(0.0)} &
\shortstack{16.5\\(0.0)} &
\shortstack{67.4\\(0.0)} &
\shortstack{40.9\\(0.0)} &
\shortstack{71.4\\(0.0)} &
\shortstack{33.5\\(0.0)} &
\shortstack{84.3\\(2.0)} &
\shortstack{11.9\\(0.2)} &
\shortstack{67.4\\(0.1)} &
\shortstack{6.1\\(0.0)}
\\
\hline


SynthSleepNet &
\makecell[l]{EEG, EMG} &
\shortstack{\textbf{74.9}\\(0.3)} &
\shortstack{\textbf{69.4}\\(0.3)} &
\shortstack{\textbf{92.7}\\(0.1)} &
\shortstack{\textbf{72.8}\\(0.2)} &
\shortstack{92.9\\(0.6)} &
\shortstack{13.3\\(0.1)} &
\shortstack{\secondbest{77.3}\\(0.2)} &
\shortstack{\secondbest{7.2}\\(0.0)} &
\shortstack{89.9\\(0.1)} &
\shortstack{50.6\\(0.1)} &
\shortstack{89.9\\(0.0)} &
\shortstack{46.7\\(0.0)} &
\shortstack{72.5\\(0.1)} &
\shortstack{36.3\\(0.1)} &
\shortstack{69.5\\(0.1)} &
\shortstack{28.4\\(0.1)} &
\shortstack{89.4\\(0.4)} &
\shortstack{14.7\\(0.3)} &
\shortstack{72.1\\(0.1)} &
\shortstack{8.1\\(0.1)} 
\\
\hline


\multirow{2}{*}{NeuroNet} &
$c_1$ &
\shortstack{67.7\\(1.6)} &
\shortstack{61.5\\(0.9)} &
\shortstack{88.5\\(0.4)} &
\shortstack{64.6\\(0.8)} &

\shortstack{89.9\\(0.9)} &
\shortstack{10.8\\(0.3)} &
\shortstack{72.5\\(0.6)} &
\shortstack{5.2\\(0.1)} &

\shortstack{88.2\\(0.8)} &
\shortstack{44.6\\(1.3)} &
\shortstack{84.8\\(0.5)} &
\shortstack{36.7\\(1.2)} &
\shortstack{50.9\\(0.5)} &
\shortstack{28.8\\(0.1)} &
\shortstack{56.2\\(0.0)} &
\shortstack{21.5\\(0.0)} &
\shortstack{89.4\\(0.2)} &
\shortstack{9.8\\(0.8)} &
\shortstack{69.0\\(1.4)} &
\shortstack{6.0\\(0.4)}
\\

\cline{2-22}

&
$c_2$ &
\shortstack{53.9\\(0.6)} &
\shortstack{49.8\\(0.3)} &
\shortstack{81.9\\(0.1)} &
\shortstack{52.1\\(0.1)} &

\shortstack{91.3\\(0.2)} &
\shortstack{13.0\\(0.2)} &
\shortstack{68.6\\(0.3)} &
\shortstack{5.8\\(0.1)} &

\shortstack{89.5\\(0.3)} &
\shortstack{47.9\\(0.2)} &
\shortstack{89.4\\(0.1)} &
\shortstack{41.7\\(0.1)} &

\shortstack{74.9\\(0.0)} &
\shortstack{48.2\\(0.0)} &
\shortstack{79.8\\(0.0)} &
\shortstack{47.1\\(0.1)} &
\shortstack{88.0\\(0.1)} &
\shortstack{20.3\\(0.0)} &
\shortstack{76.8\\(0.0)} &
\shortstack{14.8\\(0.1)}
\\

\hline
\hline

\multirow{2}{*}{\textbf{Ours}} &
\shortstack[l]{$c_1,c_2$\\(Concat)} &
\shortstack{\secondbest{72.4}\\(0.8)} &
\shortstack{\secondbest{67.3}\\(0.6)} &
\shortstack{\secondbest{91.9}\\(0.1)} &
\shortstack{\secondbest{71.4}\\(0.4)} & 
\shortstack{\secondbest{93.4}\\(0.4)} &
\shortstack{\secondbest{14.4}\\(0.4)} &
\shortstack{76.7\\(0.2)} &
\shortstack{\secondbest{7.2}\\(0.2)} &
\shortstack{\textbf{91.1}\\(0.4)} &
\shortstack{\textbf{54.2}\\(0.6)} &
\shortstack{\textbf{91.3}\\(0.2)} &
\shortstack{\secondbest{48.7}\\(0.7)} &
\shortstack{75.0\\(0.0)} &
\shortstack{47.8\\(0.0)} &
\shortstack{79.2\\(0.0)} &
\shortstack{46.3\\(0.1)} &
\shortstack{\secondbest{90.9}\\(0.5)} &
\shortstack{\secondbest{24.0}\\(0.9)} &
\shortstack{\secondbest{80.9}\\(0.1)} &
\shortstack{\secondbest{16.2}\\(0.9)}
\\
\cline{2-22}

&
\shortstack[l]{$c_1,c_2$\\(SleepPairNet)} &
\shortstack{72.1\\(0.5)} &
\shortstack{66.8\\(0.4)} &
\shortstack{91.5\\(0.1)} &
\shortstack{70.4\\(0.2)} &
\shortstack{\textbf{94.4}\\(0.1)} &
\shortstack{\textbf{18.7}\\(0.1)} &
\shortstack{\textbf{81.4}\\(0.2)} &
\shortstack{\textbf{9.9}\\(0.1)} &
\shortstack{90.1\\(0.5)} &
\shortstack{52.0\\(0.6)} &
\shortstack{\secondbest{90.8}\\(0.1)} &
\shortstack{46.6\\(0.6)} &
\shortstack{\textbf{87.9}\\(0.0)} &
\shortstack{\textbf{62.4}\\(0.1)} &
\shortstack{\textbf{87.0}\\(0.0)} &
\shortstack{\textbf{63.6}\\(0.1)} &
\shortstack{\textbf{93.4}\\(0.1)} &
\shortstack{\textbf{33.4}\\(0.2)} &
\shortstack{\textbf{85.7}\\(0.2)} &
\shortstack{\textbf{24.5}\\(0.2)}
\\



\multicolumn{2}{l|}{
\cellcolor{summaryrow}
$\Delta$F1\,/\,$\Delta$AUPRC
} &
\multicolumn{4}{c|}{
\cellcolor{summaryrow}\gnil{$-$2.1}\,/\,\gnil{$-$1.4}
} &
\multicolumn{4}{c|}{
\cellcolor{summaryrow}\gpos{5.4}\,/\,\gpos{2.7}
} &
\multicolumn{4}{c|}{
\cellcolor{summaryrow}\gpos{1.6}\,/\,\gnil{$-$0.5}
} &
\multicolumn{4}{c|}{
\cellcolor{summaryrow}\gpos{10.1}\,/\,\gpos{14.4}
} &
\multicolumn{4}{c}{
\cellcolor{summaryrow}\gpos{13.1}\,/\,\gpos{9.7}
}
\\
\hline 
\end{tabular}

\parbox{\textwidth}{
\fontsize{9}{10.0}\selectfont
\textit{Note:} Task-specific
$(c_1,c_2)$ pairs:
Sleep = (EEG C4--M1, EMG CHIN1--CHIN2);
Apnea = (EEG F3--M2, SpO$_2$);
Arousal = (EEG C4--M1, EMG CHIN1--CHIN2);
Desaturation = (CAPNO, SpO$_2$);
Hypopnea = (EEG F4--M1, SpO$_2$).} 

\end{table*}

\section{Results and Discussion}
Table~\ref{tab:nch_baseline_comparison} shows the linear-probing results on the NCH test set. Given the extreme class imbalance in event-detection tasks, we emphasize F1 and AUPRC as the primary evaluation metrics. SleepPairNet outperforms all evaluated baselines on respiratory tasks in both metrics, with the largest gains for hypopnea: 5.8 percentage points (pp) in F1 and AUPRC. For sleep scoring and arousal, SynthSleepNet and SleepFM perform best, respectively.

Table~\ref{tab:chat_comparison} evaluates transfer to CHAT by freezing the NCH-pretrained models and fitting only a linear probe. SleepPairNet again outperforms all baselines on every respiratory task, with gains larger than on NCH (up to 14.4 pp in AUPRC for desaturation). Some NCH channels are unavailable in CHAT: for apnea, EOG ROC--M1 is replaced by EEG F3--M2, and for hypopnea, EEG CZ--O1 is replaced by EEG F4--M1. Even with these substituted pairs, SleepPairNet gains 5.4 and 13.1 pp in F1,
respectively, showing that a single model transfers across both cohorts and channel combinations. For sleep scoring, its gap to SynthSleepNet narrows to 2.6 pp in F1 and 2.4 pp in AUPRC.

The Concat ablation shows that Stage-2 cross-modal pretraining adds consistent respiratory gains: SleepPairNet improves F1 over Concat by 3.6--4.8 pp on NCH and 4.3--14.6 pp on CHAT, while the two stay within 2.2 pp on sleep scoring and arousal. Concat itself is also often competitive
with existing multimodal SSL models, suggesting that the Stage-1
encoders provide a strong foundation for Stage 2.

The remaining EEG-task gap may partly reflect each model's pretraining.
SleepFM is pretrained jointly on brain activity (EEG, EOG), respiratory,
and EMG signals, while the evaluated SynthSleepNet models are pretrained
on EEG--EOG/EMG combinations, potentially favoring sleep scoring and
arousal. SleepPairNet instead samples pairs from all 16 channels, trading
repeated exposure to EEG-centric pairs for broader coverage of
physiological signals. On CHAT, however, it narrows its EEG-task gaps
and widens its respiratory gains. This suggests exposure to diverse pairs may
yield representations that transfer more robustly to new cohorts.

Random-pair training also provides flexibility across channel
combinations. Fixed-combination models such as SynthSleepNet require new
multimodal pretraining whenever the combination changes (e.g., EEG--EOG
to EEG--RESP). In contrast, a single SleepPairNet model handles all 120 pairs of its 16 channel-specific encoders without further
cross-modal pretraining, reducing compute and storage costs.

\section{Conclusion}
We introduced SleepPairNet, which learns multimodal sleep
representations through random-pair cross-attention. SleepPairNet
outperforms the evaluated baselines on respiratory tasks on NCH, with
even larger gains on the external CHAT cohort, while remaining
competitive on EEG-related tasks. Future work could compare random- and fixed-pair pretraining, evaluate on held-out
channel pairs, and extend SleepPairNet beyond pairs to larger sets of
physiological signals.

\section{Acknowledgement}
The Childhood Adenotonsillectomy Trial (CHAT)
was supported by the National Institutes of Health
(HL083075, HL083129, UL1-RR-024134, UL1 RR024989).
The National Sleep Research Resource was supported
by the National Heart, Lung, and Blood Institute (R24
HL114473, 75N92019R002). This work was partially supported by the AI Acceleration
Program at UNC Chapel Hill.

\bibliographystyle{IEEEbib}
\bibliography{strings,refs}

\end{document}